\documentclass[journal,onecolumn,12pt]{IEEEtran}
\usepackage[T1]{fontenc}
\usepackage{graphicx}
\usepackage[round]{natbib}
\begin{document}
\title{Co-Movement and Trust Development in Human-Robot Teams}
%
%
\author{Nicola Webb\textsuperscript{1}\footnote{[1] School of Engineering Mathematics and Technology, University of Bristol, UK. \\ \indent \indent \indent email: nicola.webb@bristol.ac.uk, edmund.hunt@bristol.ac.uk},
Sanja Milivojevic\textsuperscript{2}\footnote{[2] Bristol Digital Futures Institute, University of Bristol, UK}, Mehdi Sobhani\textsuperscript{1},
Zachary R. Madin\textsuperscript{1}, \\James C. Ward\textsuperscript{1}, Sagir Yusuf \textsuperscript{3}\footnote{[3] School of Computer Science, University of Birmingham, UK}, Chris Baber\textsuperscript{3} and Edmund R. Hunt\textsuperscript{1} 
}

%

%
\maketitle              
\begin{abstract}

For humans and robots to form an effective human-robot team (HRT) there must be sufficient trust between team members throughout a mission. We analyze data from an HRT experiment focused on trust dynamics in teams of one human and two robots, where trust was manipulated by robots becoming temporarily unresponsive. Whole-body movement tracking was achieved using ultrasound beacons, alongside communications and performance logs from a human-robot interface. We find evidence that synchronization between time series of human-robot movement, within a certain spatial proximity, is correlated with changes in self-reported trust. This suggests that the interplay of proxemics and kinesics --- i.e. moving together through space, where implicit communication via coordination can occur --- could play a role in building and maintaining trust in human-robot teams. Thus, quantitative indicators of coordination dynamics between team members could be used to predict trust over time and also provide early warning signals of the need for timely trust repair if trust is damaged. Hence, we aim to develop the metrology of trust in mobile human-robot teams.

\end{abstract}
\section{Introduction}
Multi-robot systems are becoming an increasingly important part of emergency service capabilities, for instance in firefighting or search and rescue operations \citep{Queralta2020}. To obtain the benefits of human-robot teaming (HRT), such as faster, safer and more effective search, team members will need to trust each other sufficiently over time in a distributed fashion \citep{Huang2021}. We have presented some of our initial thinking about trust in HRT in \citet{hunt2023steps}, where we intend to practically operationalize trust concepts by identifying a set of measurable factors that are associated with trusting relationships between individuals in HRT \citep{Hancock2020}. Rather than seeing trust as a static feature of relationships, we consider that it is likely to vary over time as team members successfully (or unsuccessfully) interact with other, positing a `ladder of trust' metaphor \citep{Baber2023}. A successful mission depends on maintaining inter-agent trust above a sufficient, context-dependent trust threshold: hence our project philosophy of `satisficing trust' \citep{hunt2023steps}. In general, trust involves taking a risk in relying on trustee, in relation to their capability, predictability or integrity (as in the model of \cite{Lewis2022}), with this risk possibly extending as far as one's own personal safety (e.g., \cite{Camara2021}). Here, we investigate the possibility of obtaining objective behavioral indicators of trust-building or maintaining processes, for example, in relation to the dynamics of inter-agent proxemics and kinesics (movement synchronization). Such an indicator could be associated with processes of nonverbal (implicit) communication between team members, which we hypothesize are an important feature of trusting teams. Therefore, we see that identifying implicit communication markers contributes to obtaining indicators of the development of trusting relationships (or their impairment). A behavioral indicator of trust could provide an autonomous system, or its supervisor, an early warning that trust is in need of repair, for which various strategies are being explored by the HRI community \citep{Baker2018}. Moreover, a period of renewed movement synchronization could itself contribute to trust repair.

In our experiment, we explore the trust-building processes of an ad hoc HRT operation, whereby participants would team up with two rover robots that they had not previously encountered to search an unstructured environment for targets. The experimental space was around 12 $\times$ 18 m in its widest extent and around 200 sq m, a larger space than usually considered in HRI trust studies. Deploying multiple (two) robots in a large space is a step toward a realistic test of the `distributed dynamic team trust' concept, e.g. of \citet{Huang2021}, whereby the level of trust between each pair of agents will vary over time. To provoke changes in trust, one condition in the experiment involved the robots becoming uncooperative through a temporary interruption to human-robot explicit communication: this was expected to induce a violation of human-robot trust. In the control condition, such an interruption did not occur. In \citet{Milivojevic2024} we examine the experimental results with a focus on participants' self-reported trust levels in questionnaires and interviews, in relation to `swift trust' and its components.  Going beyond self-reported data, we also tracked whole-body movement dynamics in the extensive test environment via an ultrasound tracking system, and this time series behavioral data is the focus here.  For the present paper, our main research question was the following: 

\textit{\textbf{RQ}: What observable behavioral changes (in movement characteristics) can be identified during an HRT mission, which might be markers of (impaired) implicit communication, and hence be used as indicators of trust development?}\\

   \begin{figure}[]

      \centering
        \includegraphics[width=0.8\columnwidth]{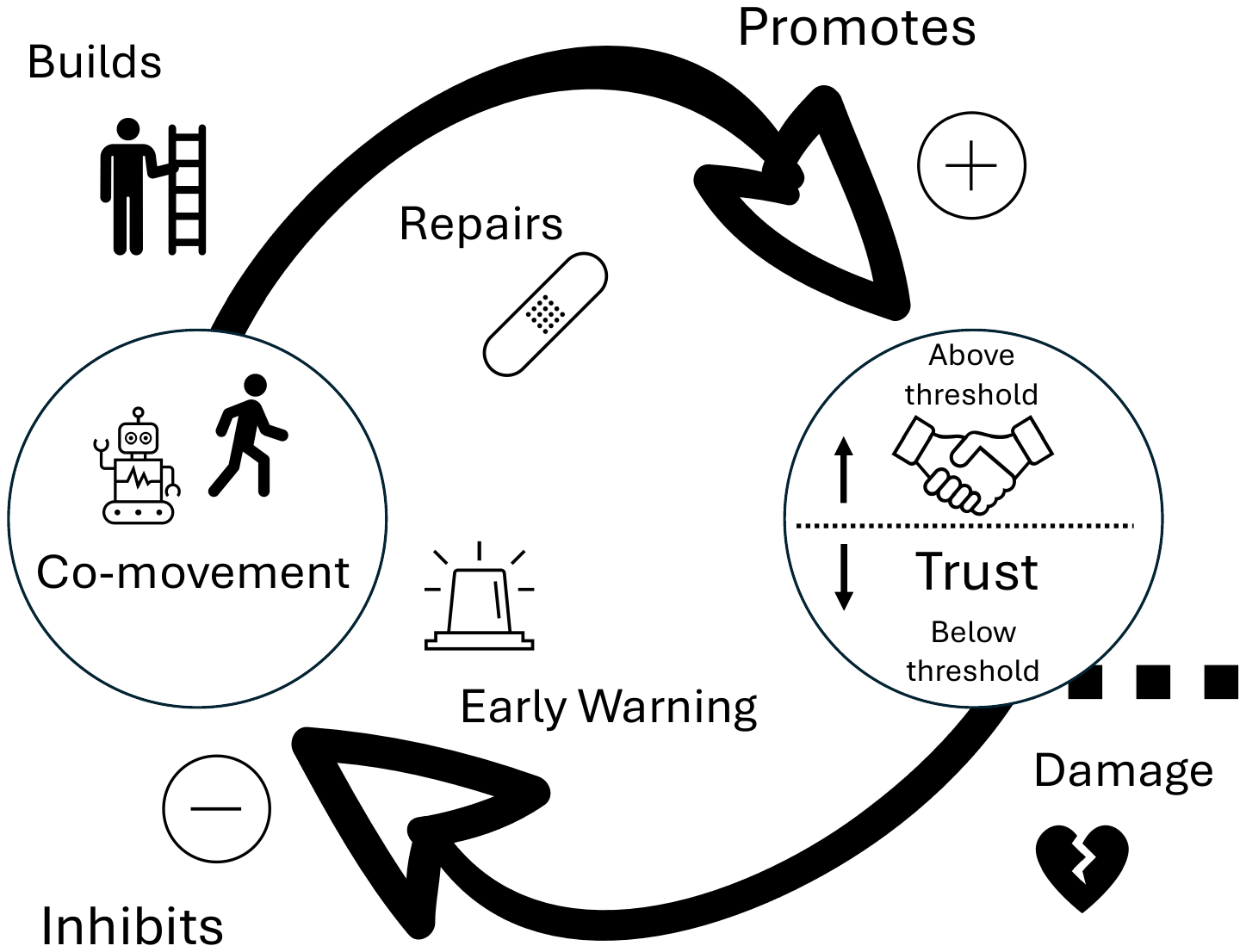}
        \label{fig:co_movement}

      \caption{The hypothesized relationships between co-movement and trust: human-robot co-movement can build and repair trust, and damaged trust below the required threshold can lead to inhibited co-movement, which therefore may be used as an early warning signal of trust impairment.}

   \end{figure}

Hypothesized relationships between human-robot co-movement and trust development (promotion and inhibition) are summarized in Figure~\ref{fig:co_movement}: we use the term `trust development' to cover both causal relationships.  We briefly present related work on nonverbal communication and swift trust (Section 2), before detailing our experimental methodology (Section 3) and outlining findings (Section 4). We then provide ideas for future work (Section 5). 

\section{Related Work}

Nonverbal communication is a vital part of human-human communication \citep{Urakami2023}, and thus understanding the range of nonverbal cues that might be afforded, by design or incidentally, by robot behavior is an important opportunity for HRI. In particular, proxemics and kinesics are important modalities for human-robot interaction \citep{Saunderson2019}. It is well known that people have preferred social distances from others during interaction and that proximity shapes engagement \citep{Saunderson2019}. \cite{peters2018investigating} found that people responded similarly to both virtual human and robot characters in their choice of social separation distance (1.23 m on average). Moreover, the way a robot moves can convey implicit motion cues to a user -- humans quickly perceive apparent goal-directed behavior from motion \citep{vanBuren2016}. In view of this, some recent HRI research has focused on motion planning for robots that is `legible' with respect to conveying cues of the robot's intent, leading to more fluent collaboration \citep{Dragan2015,lichtenthaler2016}.

\citet{Cha2018} surveyed nonverbal signalling methods for non-humanoid robots, which is relevant to the present study involving rovers. Their goal was to ``support varying levels of interaction while increasing transparency of a robot's internal state'' \citep{Cha2018}. They identified different levels of interaction, starting with mere `coexistence' (where humans are bystanders or observers), rising to `coordination' (coordinating actions in time or space, for greater efficiency or conflict prevention), and then describing the highest level of interaction as `collaboration' (actively working toward a shared goal by communicating such that actions are complementary). The present study, involving two rovers that purposefully navigate around the perimeter of the experimental space until the user calls them to help, has all three levels of interaction. Users can watch the robots as they `autonomously' search the room (coexistence), walk alongside or nearby a robot with mutual care between the human and robot not to collide with each other (coordination), and also stand alongside a robot for a simulated joint search task (collaboration). We anticipate that each of these three increasing levels of interaction are likely to have an increasing impact on trust levels, for better or worse depending if the interaction is successful; and that there is likely to be a relation between decreasing social distance and these higher levels of interaction occurring.

During the robots' movement throughout the experimental space, and when they are called to perform a joint inspection with the user, the `autonomy' (we use a `Wizard-of-Oz' protocol) is sufficient to travel without collision with the environment or the user, including when the user walks alongside the robot. This is an example of emergent coordination in joint action \citep{Knoblich2011}, and we hypothesize that this tends to improve a trusting relationship between the user and their `teammate'. In the experiment, described in the next section, participants were naive members of the public who had mostly never interacted with robots before. In this context, users will form a provisional level of trust in the robot system: this has been referred to as `swift trust' \citep{Haring2021}. With swift trust, users are reliant on `imported information' (prior knowledge relevant to the context, e.g. the university running the trial may be perceived to be trustworthy), surface-level cues (for example, the appearance of the robot), and the user's propensity to trust \citep{Haring2021}. Given this provisional trust, we anticipate trust levels to show significant plasticity -- and for the nonverbal interaction via kinesics and proxemics to have an important influence on trust development. 

A link between proxemics and trust has been proposed in the `Physical Trust Requirement' model of  \cite{Camara2021,Camara2022}. This was initially intended for the specific case of pedestrian interaction with autonomous vehicles (cars) where pedestrians will be concerned for their safety crossing the road. In follow-up work \citep{Camara2022} this was extended to model human-human and human-humanoid interaction. In either case, the `trust zone' is the region whereby one agent would be reliant on the other to slow down in time to avoid collision. Thus, a higher robot speed implies a bigger trust zone around it where a human would need it to reliably slow down if it were on a collision course. For the `Leo Rover' robots used in this study, they have a maximum linear speed of $\sim$0.4 ms$^{-1}$, which is faster than the NAO robot (0.3 ms$^{-1}$) but slower than the PR2 robot (1 ms$^{-1}$) considered in \cite{Camara2022}. Depending on the human and `robot' reaction times, coefficient of friction, approach angle, and whether one or both agents are distracted, this will imply trust regions of varying sizes, notwithstanding that the Leo Rovers do not pose a meaningful physical collision risk (30 cm tall, 6.5 kg mass). All considered we choose 2 m as a plausible vicinity within which trust will be developed, both in relation to task coordination and collaboration (joint action), and to collision avoidance. 

\section{Experiment}
\begin{figure}[t]

\begin{minipage}[c]{0.35\linewidth}

      \centering
      \includegraphics[width=1\columnwidth]{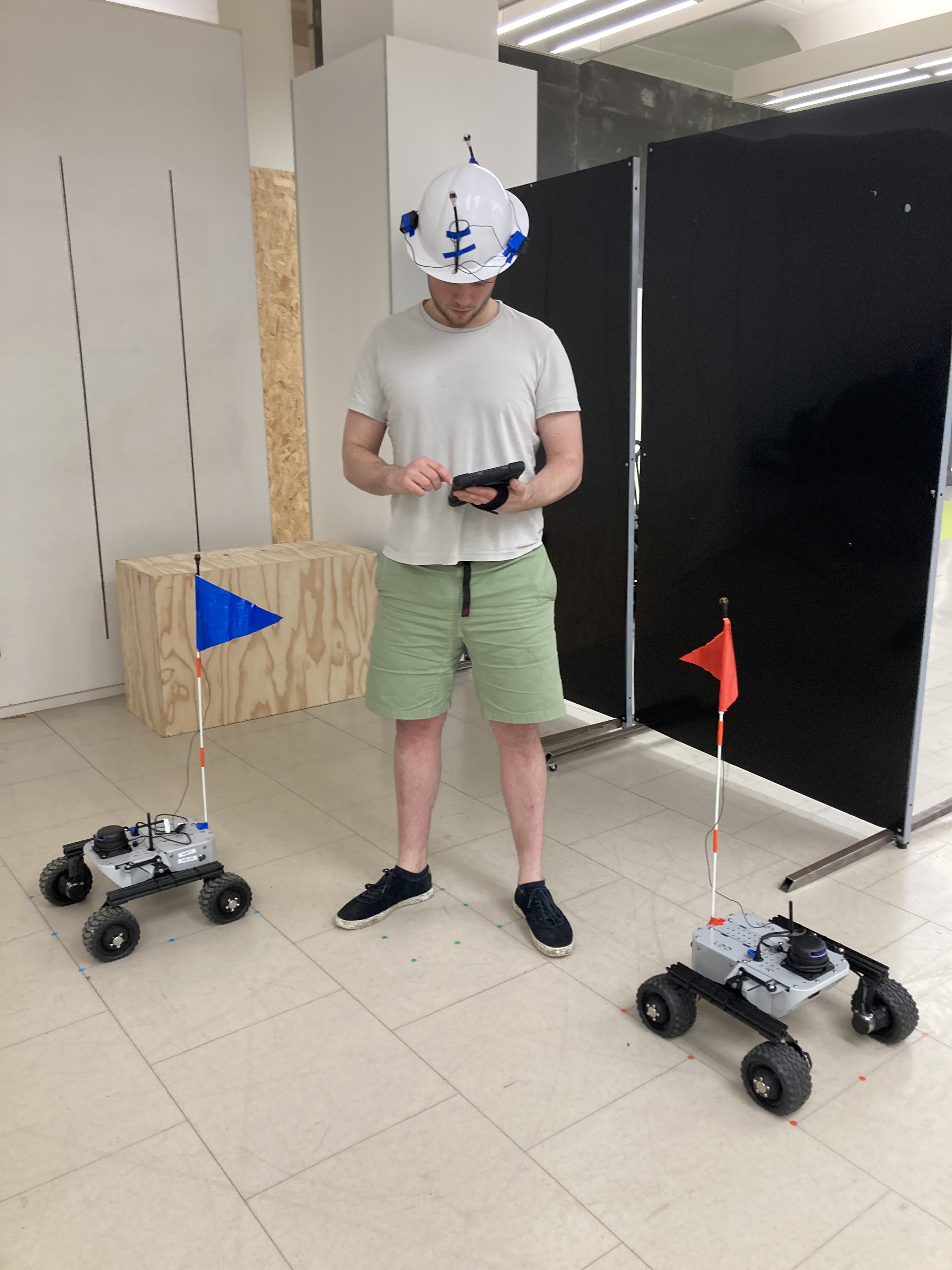}    
      \caption{Participant and robot starting locations. Participants wore a white hard hat with ultrasound tracking microphones, which were also on the robots.}
      \label{fig:hat_interface}
\end{minipage}
\hfill
\begin{minipage}[c]{0.55\linewidth}
  \centering
      \includegraphics[width=1\columnwidth]{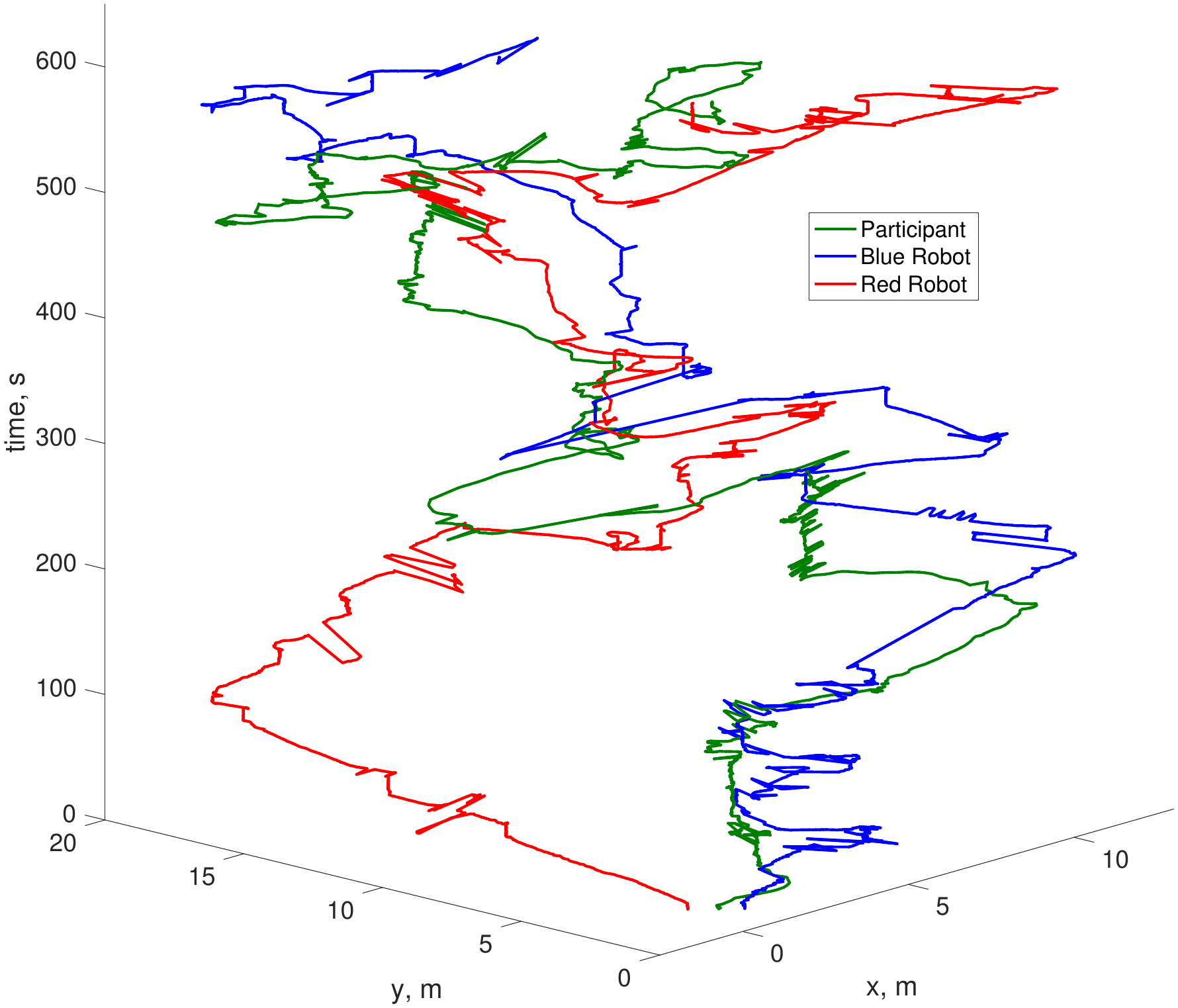}
      \caption{An example of the indoor positioning system tracking trajectories. The z-axis shows time (this trial lasted 10 m 42 s). Most participants (green line) searched around the space counter-clockwise, which favoured initial coordination and collaboration with the blue robot.}
      \label{fig:trajectory}
      \end{minipage}

   \end{figure}

We carried out an HRI experiment on 4 consecutive days (11–14 December 2023) with 22 participants. The participants worked with two four-wheeled rover robots, shown in Figure \ref{fig:hat_interface}, to collaborate on a search task in an environment. The robots were remotely operated by two experimenters (i.e. a ‘Wizard-of-Oz’ protocol) such that messages from the participant to the robot would go to the Wizards who would respond with the relevant behavior (driving, ‘scanning’ and messaging). The search task involved checking the imagined gas pipework by scanning waist-height QR codes with a handheld tablet (Figure \ref{fig:hat_interface}). The human-robot tablet interface had buttons for the user to select the `red' or `blue' robot (indicated by their flag) and press `I need help' to call it for the joint inspection task; and a `Camera' button for the QR code scanning. Sometimes the QR codes indicated `damaged' pipework that required collaboration with a robot teammate to inspect. The participant had to call a robot and wait while the robot pointed its front-facing camera at the co-located robot-height QR code. In the absence of such an instruction, the robots travelled around the outside of the experimental space `autonomously' scanning robot-height QR codes (ten in total), performing one lap of the space before returning to the start position. There were five QR codes for the participant to scan as an individual task (without the robot), and five to be done as a joint task. 

\subsection{Experimental conditions}

For half of the participants, the communications were relayed without problem between the human and robots (‘\textit{Full communications}’, `FC' condition). For the other half, communications were lost for 3 minutes during the experiment, from \textit{t} = 3 m to \textit{t} = 6 m (‘\textit{Interrupted communications}’, `IC' condition). Robots would send an initial message (`Possible communications problem'), and afterwards `Robot unavailable', and not respond when called. This meant participants were unable to scan the collaborative QR codes during this time -- but after 6 minutes (when robots sent a single message `Communications restored') they could return to perform the joint scan tasks if they chose to. Clearly, the robots being unresponsive will impair collaboration during the interruption, and therefore we primarily focus on the participant behavior after communications have been restored.

\subsection{Data collection}
We collected quantitative and qualitative data to assess trust changes throughout each experimental trial. We focus on data relating to movement in this paper. After completing a consent form, participants answered some basic demographic questions: there were $N=22$ participants, 10 men and 12 women, with a median age range of 25--34, and median education of a bachelor's degree. On prior robotics experience, most (15) answered `none' and the rest (7) answered `limited'. Participants were given a briefing on the task and the operation of the robot system. This included a hands-on demonstration of the interface where participants would have to scan one individual QR code and one requiring them to call a robot for help. Following this, participants completed a pre-trial trust survey, the 14-item `trust perception scale' of \cite{Schaefer2016}. After the experimental trial, participants completed an additional, post-interaction trust survey and then were interviewed on their experiences.

During the experimental trial, we logged every time the participant scanned a QR Code and every time a message was sent by the participant or robot (Wizard). We also recorded tracking data (Figure \ref{fig:trajectory}) on the movement of the participant and the two robots using an indoor ultrasound positioning system (\textit{Marvelmind Robotics}). The participant was asked to wear a white hard hat, resembling the helmet a firefighter might wear, with two tracking beacons on the top (front and back, Figure \ref{fig:hat_interface}). Due to the unexpected data loss of one trajectory from each condition, we analyse a total of \textit{N} = 20 sets of movement trajectories. 

Using the tracking data, we examine inter-agent separation dynamics (proxemics), whereby a closer proximity could indicate a closer working relationship, and possibly, greater implicit communication and trust. We also calculate the cross-approximate entropy (‘XApEn’) between the time series of movement speeds (kinesics) of participants and the two robots. Approximate entropy can be used to quantify the amount of regularity (or unpredictability) in time series data \citep{Pincus1991}. The cross-approximate entropy is used here to quantify the similarity in movement patterns (kinesics) between the human and the robot. We hypothesize that a closer working relationship with a robot (favouring development of trust) is associated with lower XApEn (a higher synchronization in movement speeds). This could be indicative of nonverbal (implicit) communication processes, i.e. coordination of movement.

\section{Data Analysis}

\subsection{Trust change}

\begin{figure}[h]
\begin{minipage}[c]{0.50\linewidth}

      \centering
      \includegraphics[width=1\columnwidth]{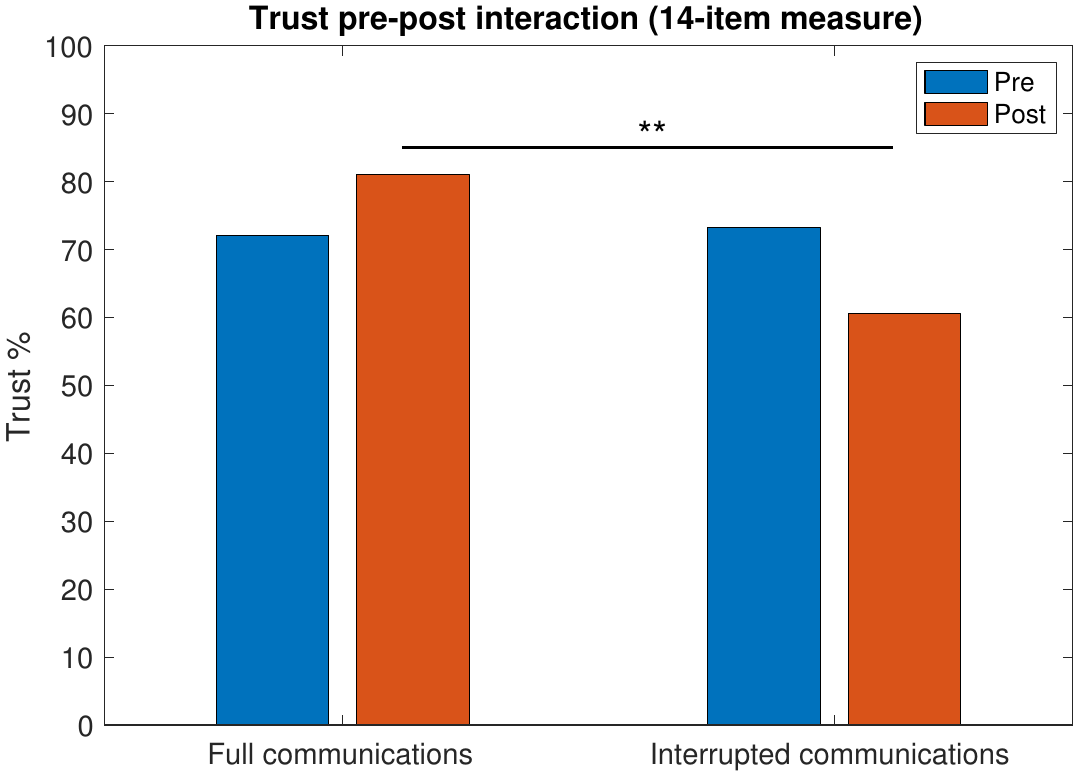}
      \caption{Participant mean trust pre- and post-interaction by condition.}
      \label{fig:trust_survey}

\end{minipage}
\hfill
\begin{minipage}[c]{0.45\linewidth}
      \centering
      \includegraphics[width=1\columnwidth]{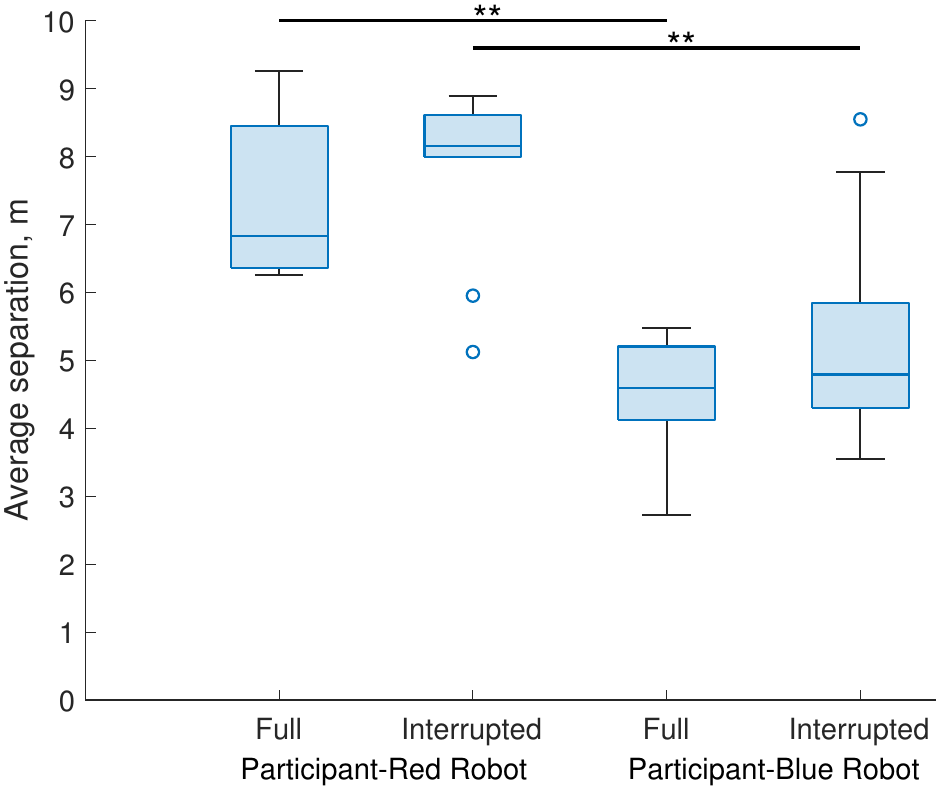}    
      \caption{Participants were significantly closer to the blue robot during trials, in both conditions.}
      \label{fig:proxemics}

\end{minipage}
\vspace{-1.2cm}
\end{figure}

Mean reported trust improved in the full communications condition but decreased following an interruption to communications (Fig.~\ref{fig:trust_survey}). A Kruskal-Wallis test was performed on the pre and post-interaction trust ratings in the two conditions, and there was a significant difference between the rank totals of the four groups, $H(3, n=44)=10.94, p<0.05$.  Post hoc comparisons with Tukey's HSD procedure indicated significantly lower post-interaction trust in the the `IC' condition (60.5\% on a 0--100\% scale) compared to the `FC' condition (81.1\%, rank difference: $-17.91, p<0.01$). Although the average post-interaction trust is lower in the IC condition than it was pre-interaction, this drop does not reach statistical significance (60.5\% vs. 73.3\%, p=0.22).

\subsection{Participant proxemics}
Participants were significantly closer to the blue robot in both conditions (Fig.~\ref{fig:proxemics}, $H(3, n=40)=22.75, p<0.001$, Tukey's HSD probabilities both $p<0.01$). This is apparently because they tended to begin their search in an anticlockwise direction (see, e.g., Fig.~\ref{fig:trajectory}). We assume this is because, during the hands-on demonstration of the QR Code scanning, the two example locations were located in an anti-clockwise direction from the starting position, and participants would typically begin their trial by re-scanning the same locations.

\subsection{Analysis of Communication and Performance Logs}

We performed a Two-way Mixed Analysis of Variance (\textit{Condition} as Between Subject and \textit{Robots} as Within Subject factors) for Communications (messages sent) and Tasks Completed with Robot. There was a significant main effect of Condition on Communications [$F(1,21) = 89.96, p<0.001$, observed power = 1] but no effect of Robot.  This indicates that more messages were sent in the `IC' condition (18.1 messages to Blue robot, 17.3 messages to Red robot) than in the `FC' condition (7.3 messages to Blue, 5.1 messages to Red).  

There was a significant main effect of Robot on Tasks Completed with Robot [$F(1,21) = 6.124, p<0.05$, observed power 0.656] but no effect of Condition and no interaction effects.  This suggests that participants tended to complete more tasks (shared QR code scans) with the Blue robot (2.7 in the `FC' condition, 3.2 in the `IC' condition) than the Red robot (2.2 in the `FC' condition, 1.2 in the `IC' condition).

In terms of individual tasks, there was a significant difference between Conditions ($t(21) = 2.278, p<0.05$), with participants completing more individual scans in the `IC' condition (6.4) than the `FC' condition (5.3).  

 The `IC' condition was divided into three periods for the cross-entropy analysis: 0-3 m (pre-interruption), 3-6 m (during the interruption), and after 6 m, where communications were restored and collaboration with the robot became possible again. Participants spent significantly more time completing the `IC' condition ($t(21) = 2.537, p<0.05$) than the `FC' condition (8 minutes 28 seconds for the `IC' condition versus 6 minutes 15 seconds for the `FC' condition). This meant that the post 6 m period allowed for a meaningful time for the behavioral impact of the communications disruption to become evident. Because the `FC' condition experienced no such disruption, we divided data for that condition into two approximately equal parts (before and after 3 minutes), allowing direct comparison between conditions for the initial 0-3 m period.

   \begin{figure}[t]
      \centering
      \includegraphics[width=1\columnwidth]{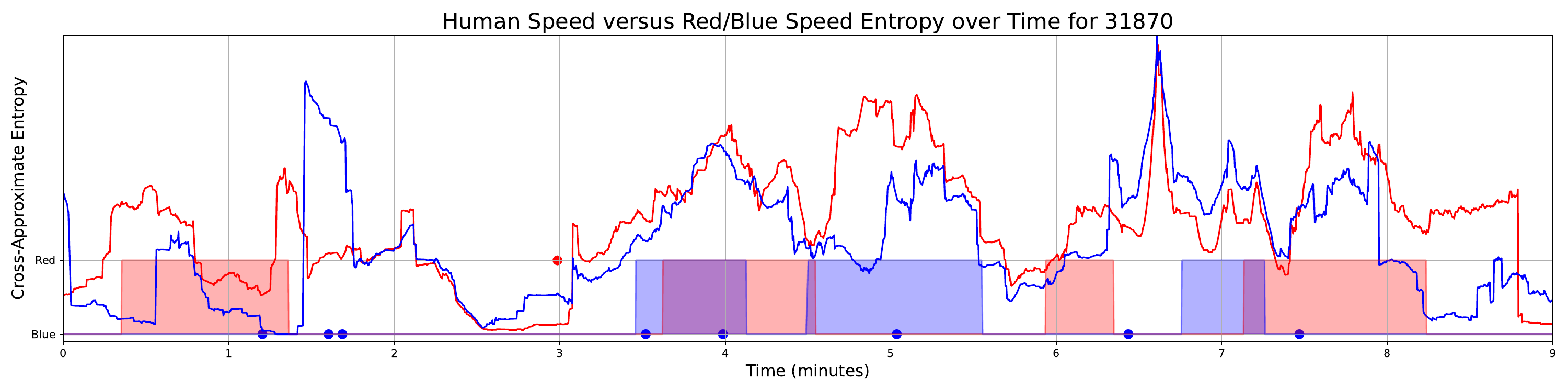}
        \includegraphics[width=1\columnwidth]{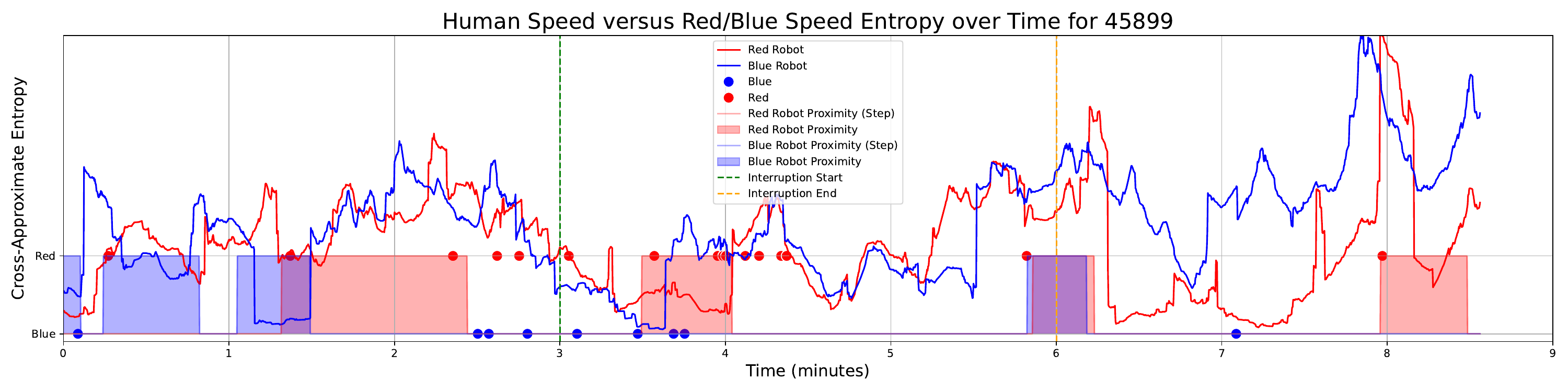}
      \caption{An example of cross-approximate entropy over time for participants (ID in title) in the `FC' (top) and `IC' conditions (bottom). Shaded red and blue periods indicate the participant and relevant robot within 2 m proximity. The series of call events (red or blue) are shown by dots. In the IC condition, during the 3-6 m time period, indicated by dotted vertical lines, the robots would not respond to calls.}
      \label{fig:time_series}
      \vspace{-0.3cm}
   \end{figure}

   \begin{figure}[t]
      \centering
      \includegraphics[width=1\columnwidth]{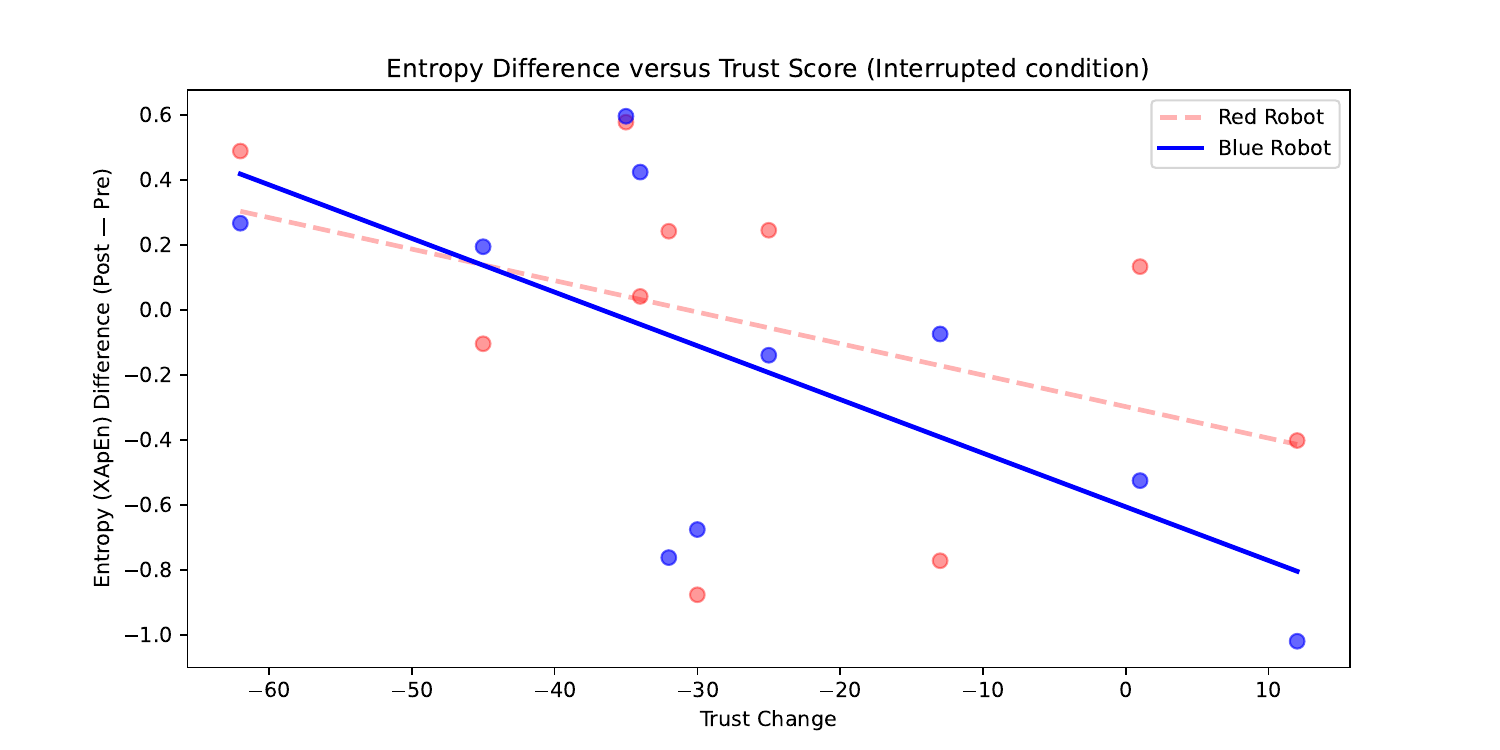}
      \caption{The change in cross-approximate entropy between the participants and the red and blue robots, before $t=3$ m and after $t=6$ m, correlated with the change in trust.}
      \vspace{-0.3cm}
      \label{fig:entropy_change}
   \end{figure}

\subsection{Movement synchronization with proximate robots}
We analyzed the periods of movement when a robot was near the participant. Specifically, we considered periods of time where the proximity was within 2 metres, which is generally considered to be within a person's social space \citep{cristani2011social,peters2018investigating}. Figure \ref{fig:time_series} shows examples of the cross-entropy over time from each of the two conditions (full and interrupted communications). Figure \ref{fig:entropy_change} shows the individual-level change in movement synchronisation with each robot (red and blue) versus that individual's overall change in trust after the interruption to explicit communications. That is, it shows the difference in cross-approximate entropy, `XApEn', after and before the period of interruption (i.e., after 6m and before 3 m) and the change in the trust questionnaire score \citep{Schaefer2016}, before and after the experiment. In the `IC' condition, there are negative correlations with both the red and blue robots (Pearson correlations: $-0.42, p=0.23$, $-0.65, p=0.04$), the correlation with the blue robot meeting a 5\% confidence level of statistical significance. That is to say, where there was a greater decrease in synchronisation (\textit{higher} XApEN) of human and robot movement speeds, this was associated with a greater loss of trust reported post-experiment. Interestingly, a minority of participants report little change in overall trust (e.g., $\pm 10\%$), and these show mainly a decrease in cross-entropy. Given that most participants interacted more with the blue robot (e.g. Fig. \ref{fig:proxemics}) a stronger effect on post-interruption human-blue robot coordination might indeed be anticipated. Considering that participants had an opportunity to work with two robots (red or blue), in future studies it might be preferable to identify a change in trust toward each robot specifically, as well as the robots in general: i.e. to distinguish `component-specific' as well as `system-wide' trust \citep{Keller2009}. In the full communications condition, there were not any significant correlations between entropy change and trust change (red and blue: $\rho=0.18, p=0.61$; $\rho=0.34, p=0.34$). In this case, we suggest that because participants faced constant conditions of full communications over time, there are unlikely to be clear changes in human-robot movement synchronization.

\subsection{Selected interview findings}
Post-experiment interviews were wide-ranging regarding the participants' experiences: here we report a relevant selection of remarks pertaining to proxemics and kinesics, and their relation to trust (participant ID in brackets). Regarding robot preference –- which robot did they choose to call when needed a collaborative human-robot reading of QR codes -- most participants indicated the proximity of the robot as the critical factor in decision-making (14 out of 22). The perception of the red robot was that the robot had just ``gone off to the left, and I was going the same sort of direction as the blue. So, I initially relied on a blue one'' (41687). Another participant mentioned that blue was their favorite, because it was ``closest. He followed me the same way around, whereas red went the other way around'' (44870; also 8223). The proximity of the robot made them a ``part of the team... something that I could use to get the job done'' (8223). Robot proximity also influenced the development of trust. The robot that was close to the human was considered to be a ‘part of the team’: ``But in that moment there I was like, yeah, OK, at least it's right next to me. It did have a positive impact. I guess... like a teammate right there. I can't reach that low, so it's going to help me and it's right there it's going to scan so, very good'' (8223). ``From the get go, red robot completely went off on his own on mission somewhere else. Well, I found more trust and I guess dependability with the blue robot to begin with. Because it was following me, yeah. Working as a part of the team.'' (41770) 

The robots’ performance was often linked to their speed. Most participants commented they were ‘quite slow’ (12369). Speed was at times a decisive factor in choosing which robot to call (25742; 41687; 45899). This is despite the fact that the robots were identical (or near-identical), yet they were perceived differently. As one participant observed, ``Yeah, I think there were a couple of times where the blue one was responsive, but it was almost like half the speed of the red one so I was like, looking at the red one like, oooh, should have gone with the red one'' (25742). Speed, especially combined with how the perceived task should be completed, often led participants to single out a favorite robot: ``Yeah, red was one that was better than the other. So, the blue one was less, maybe it still reliable, but it was a bit slow and then at points it seemed like it was moving, not, it was moving around the space in a way that wasn’t completing the mission'' (25742). Interestingly, human’s perception of robots’ contribution to the task and overall mission varied. Some respondents, for example, felt they were doing more work than the robots -- because the robots were slower (12915) or because humans were doing ``the first sweep'' (23490).

For participants who experienced the IC condition, if a robot was nearby when it gave its `unavailable' message, this was an aggravating factor: ``I had to wait for the red robot to come around, which made me think actually, I’ve got a [blue] robot right next to me... why can’t it come and help me?'' (17465; also 48619). ``Red started in this bit, so I didn't use red at all, and then I used red once and it said no... And then I tried blue because blue had been good to me. And then blue also was saying no and it's... a bit frustrating because I was, I was only using the one that was closest to me before and then when I was closest to red I tried using red and it didn't work...'' (23490). Once communications had been restored, some participants felt if robots responded to their call, they could trust them again (27837). The level of trust, however, was not the same as at the beginning of the experiment: ``I would probably... I still wouldn't go back to full trust, but I'd go back to like, half trust, because I know that [the robot] has the ability to suddenly cut out for a large period of time'' (37456).

\section{Discussion}
We carried out a human-robot teaming experiment, tracking the movement of one human and two robots in a large experimental space (around 200 sq m). We found some evidence of proxemics and kinesics relating to change in trust: there was a significant negative correlation between the cross-approximate entropy change between the movement speed of participants and the commonly-chosen `blue' robot, and change in self-reported trust (considered for times when they were within 2 m of each other). That is to say, in the experimental condition where there was an interruption to explicit communications, there was potential damage to subsequent implicit communication (movement synchronization), even though communications had been restored. A greater increase in entropy (desynchronization) was associated with a bigger decrease in trust. This could be because users became more wary of the robot and less fluid in coordinating movement (trust damage leading to desynchronization, Figure~\ref{fig:co_movement}, bottom). Qualitative interviews with participants confirmed that robot proximity was a decisive factor in which robot they preferred to call, and that a robot staying nearby over time made it ``part of the team'' (Figure~\ref{fig:co_movement}, top). Conversely, when communications were interrupted, the robot being nearby and nevertheless being unresponsive was reported to aggravate feelings of frustration.
\paragraph*{\textbf{Limitations}}
Although we identified a significant correlation between human-robot movement dynamics and trust change, we acknowledge various possible limitations in this finding that should be considered in future research. As we hypothesize in Figure~\ref{fig:co_movement}, co-movement could be used as an early warning signal of impaired trust in a human-robot team. Establishing a causal relationship from decreased trust to decreased co-movement will require further experimentation and careful analysis, beyond our initial findings here. For example, it could be that after a 3-minute interruption to robot co-working it is more difficult for the participant to find the rhythm of co-moving with the robot, which may not be a trust issue as such. While some interview findings provide evidence for a positive trust building effect of robot proximity and co-movement (``working as part of the team''), more evidence is needed to establish causal relationships.

While the post-interaction trust level was significantly lower in the `IC' condition compared to the non-interrupted condition (Figure~\ref{fig:trust_survey}), the within-condition drop (from 73.3\% average to 60.5\%) was not itself statistically significant. Nevertheless, as shown by Figure~\ref{fig:entropy_change}, at an individual level there was substantial individual variation in trust changes, with some participants showing marked decreases in trust and a few showing little change. A larger sample size may confirm a systematically effective trust manipulation while revealing the importance of individual differences (such as prior experience and propensity to trust) in trust dynamics. It is also possible that in the `IC' condition, that participants could lose trust in the communications channel itself rather attributing non-compliance to the robots. Further questioning could further explore possible distinctions between change in trust in the robots and their communication channel, which might explain part of the variation in trust change.

Finally, in this study we used two robots to construct a multi-robot HRI scenario instead of a dyadic one. However, there was no deliberate control or manipulation over the two robots to make them different in design, for example, to then compare the results. A future experiment could have one or more further conditions to examine distinct effects of multi-robot compared to single robot teaming. Nevertheless, providing two robot teammates did reveal findings such as spatial proximity (co-movement) being decisive in robot interaction choices and perceptions.

\paragraph*{\textbf{Future Work}}
Nonverbal or implicit communication is important for increasing the transparency and understandability of a robot's internal state; for improving task performance; and for recovering from errors quickly \citep{Breazeal2005}. Even though this experiment afforded only fairly minimal implicit communication (maintenance of proxemics and a degree of synchronization of kinesics), this was correlated with measured changes in trust. Following the exploratory analysis presented here, we will conduct further research on time series of reported trust to see if maintenance (decrement) in such implicit communication is indeed a robust indicator of trust building (damage). We will also consider alternative, non-intrusive measures of trust without the possible fatigue and biases of self-reporting trust on a scale: for example, by asking the participants to explicitly choose to delegate a sub-task to the robot or to perform it themselves, thus using robot reliance as a proxy trust measure. Beyond kinesics, we may also explore other forms of minimal, nonverbal communication suited to non-humanoid robots \citep{Cha2018}, such as indicator lights or sounds \citep{Cha2016}. These could be used to increase the legibility of the robots' state and mitigate the trust damage that can occur in the absence of explanation when human-robot teaming dynamics are disrupted.
\\
\paragraph*{\textbf{Acknowledgements}} 
We thank Sparks Bristol, UK, for hosting the experiment. We acknowledge funding from EPSRC (EP/X028569/1, `Satisificing Trust in Human-Robot Teams'). \\

%
%
%
%
\bibliographystyle{plainnat}
\bibliography{references}

\end{document}